# Deep Q-Network Based Multi-agent Reinforcement Learning with Binary Action Agents

**Abdul Mueed Hafiz*[1], Ghulam Mohiuddin Bhat[2]**

[1, 2] Department of Electronics and Communication Engineering
Institute of Technology, University of Kashmir
Srinagar, J&K, India, 190006.

*[1]Corresponding Author Email: mueedhafiz@uok.edu.in
[2]Co-author Email: drgmbhat@uok.edu.in

ORC-ID[1]: 0000-0002-2266-3708
ORC-ID[2]: 0000-0001-9106-4699

**Abstract**

Deep Q-Network (DQN) based multi-agent systems (MAS) for reinforcement learning (RL) use various schemes where in the agents have to learn and communicate. The learning is however specific to each agent and communication may be satisfactorily designed for the agents. As more complex Deep Q-Networks come to the fore, the overall complexity of the multi-agent system increases leading to issues like difficulty in training, need for higher resources and more training time, difficulty in fine-tuning, etc. To address these issues we propose a simple but efficient DQN based MAS for RL which uses shared state and rewards, but agent-specific actions, for updation of the experience replay pool of the DQNs, where each agent is a DQN. The benefits of the approach are overall simplicity, faster convergence and better performance as compared to conventional DQN based approaches. It should be noted that the method can be extended to any DQN. As such we use simple DQN and DDQN (Double Q-learning) respectively on three separate tasks i.e. *Cartpole-v1* (OpenAI Gym environment) , *LunarLander-v2* (OpenAI Gym environment) and *Maze Traversal* (customized environment). The proposed approach outperforms the baseline on these tasks by decent margins respectively. Code is available at: *https://github.com/mueedhafiz1982/dqn_mas_rl_bin_aa.git*

## 1. Introduction

Deep Q-Network (DQN) [1,2] based Multi-agent systems (MAS) [3,4] for Reinforcement Learning (RL) [5-13] are being researched intensely. In MAS, the emphasis is put on data-sharing schemes, inter-agent communication, and novelty of the Q-Network. However with newer systems on the horizon, the complexity is increasing leading to issues like difficulty in training, need for more resources and more time for training, fine-tuning issues, etc. To address these problems, we propose a simple but efficient DQN based MAS for RL with shared state and reward, but agent specific action updation in the experience replay pools of the DQNs, which is a first to the best of our knowledge. The benefits of the approach are overall simplicity, faster convergence and better performance than the baseline which consists of a single DQN, here simple DQN [1] and DDQN (Double Q-learning) [2]. Since the approach can be extended to any type of DQN, hence to keep amount of experimentation in perspective, we use two types of DQNs. Also, we do not compare the performance of the proposed approach with MAS for RL, because our approach is much simpler and different as compared to MAS for RL, and is not a likely extension of the contemporary approaches in this area. We use two OpenAI Gym environments i.e. CartPole-v1 and LunarLander-v2, and one customized environment i.e. Maze Traversal, in experimentation. The proposed approach outperforms the baseline on these tasks.

The rest of the paper is organized as follows. Section 2 gives the background. Section 3 gives the implementation and experimentation. We conclude in Section 4.

## 2. Background

### 2.1 Deep Q Networks

Minh et al [1] used a deep neural network for approximation of the optimal action-value function as

$$Q^*(s,a) = \max_\pi \mathbb{E}\left[\sum_{s=0}^{\infty} \gamma^s r_{t+s} | s_t = s, a_t = a, \pi\right],$$

where the expression gives the maximum sum of rewards $r_t$ using discounting factor $\gamma$ for each time step $t$ which is achieved by a behavior policy $\pi = P(a|s)$, for state $s$ and action $a$ after making an observation of the same.

Before the above work was published, it was well known that conventional reinforcement learning (RL) algorithms would become unstable or even show divergence when non linear functions were used e.g. in the case of neural network being used for representation of $Q$ i.e. the action-value function. Minh et al [1] discuss some notable reasons. Firstly, it is due to the correlations present in the observation sequence of the state $s$. In RL applications there is a sure auto-correlation in the sequence state observations which is a time-series. However certainly this could be true also of the applications of deep networks for modeling of time series data. Secondly, small updates to $Q$ can significantly alter the policy $\pi$, and can hence alter the distribution of data. Finally, due to correlations between the action values, $Q$ and the values of the targets i.e. $r + \gamma \max_{a'} Q(s',a')$

In their work, the authors address these issues by using the following. Firstly, they use a biologically inspired process which they refer to as *experience replay*. It randomizes data hence removing correlations in the observation sequence of the state $s$, and also smoothes any changes in the distribution of data. Secondly, they use an iteration based update rule which adjusts the action values, $Q$, towards the target values, $Q'$, which are only updated periodically hence reducing target correlations.

As it turns out, there can be many techniques of approximation of the action value function, $Q(s,a)$, by using a deep network. The sole input to the DQN is the state representation, and also its output layer has a different output for every action. Each output unit corresponds to the predicted $Q$-values of the separate action in the input state. The input of their DQN consists of an image (84 x 84 x 4) which is produced by using a preprocessing map $\phi$. The DQN has 4 hidden layers out of which three are Convolutional and the last one is fully-connected (i.e. dense). All the layers use a ReLU activation function. The output layer of the DQN is also a fully-connected layer with one output for every action. The $Q$-learning iterative updation of the DQN uses the loss function as

$$\mathscr{L}_i(\theta_i) = \mathbb{E}_{(s,a,r,s')\sim U(D)}\left[\left(r + \gamma \max_{a'} Q(s',a';\theta_i^-) - Q(s,a;\theta_i)\right)^2\right],$$

where $\gamma$ is the discount factor in the horizon of the agent, $\theta_i$ comprises of the parameters of the DQN for the $i^{th}$ iteration, and $\theta_i^-$ comprises of the parameters of the DQN for target computation for $i^{th}$ iteration. $\theta_i^-$, the network parameters of the target are updated with the DQN parameters $\theta_i$ for every $C$ steps and are kept fixed between updates.

In order to perform *experience replay*, the agent's experiences $e_t$ are stored which are represented by a tuple:

$$e_t = (s_t, a_t, r_t, s_{t+1})$$

which consists of the observed state, $s_t$, in the period $t$, the reward received, $r_t$, by the agent in the period $t$, the action taken, $a_t$, in the period $t$, and the resulting state, $s_{t+1}$, in period $t+1$. This stored dataset of the experiences of the agent at the period $t$ consists of the past experiences.

$$D_t = [e_1, e_2, \ldots, e_t]$$

During DQN learning, $Q$-learning updation is computed based on samples/minibatches of the experience *(s, a, r, s')* which are drawn uniformly and randomly from the collection of the stored samples $D_t$.

## 2.2 Double Q-learning

The max operator used in Deep Q-Networks, uses the same values both for selection and for evaluation an action. This leads to it being more inclined towards selection of overestimated values. This issue results in overoptimistic estimation of values. In order to prevent this, Hasselt et al [2] decoupled the selection component from the evaluation component. This is the thinking behind Double Q-learning. In the basic Double Q-learning technique, two functions are learned by assigning every experience randomly for updation of any one of these two value functions, in a manner that that there will be two sets of weights, $\theta$ and $\theta'$. For every update, one weight-set is used for the greedy policy determination and the other is used for its value determination. Decoupling the selection component and the evaluation component in Q-learning and rewriting its target, we have

$$Y_t^Q \equiv R_{t+1} + \gamma Q\left(S_{t+1}, \operatorname*{argmax}_a Q(S_{t+1}, a; \boldsymbol{\theta}_t); \boldsymbol{\theta}_t\right).$$

The error of the Double Q-learning algorithm can now be written as

$$Y_t^{DoubleQ} \equiv R_{t+1} + \gamma Q\left(S_{t+1}, \operatorname*{argmax}_a Q(S_{t+1}, a; \boldsymbol{\theta}_t); \boldsymbol{\theta}_t'\right).$$

In the above equation, the second set of weights $\boldsymbol{\theta}_t'$ is used to evaluate the value of the policy. The second set of weights can be symmetrically updated by switching the roles of $\boldsymbol{\theta}$ and $\boldsymbol{\theta}'$.

## 2.3 Multi-agent Systems

Multi-agent Reinforcement Learning (RL) is a research area which has a rich history [3,4,14]. The authors of [15] identified modularity as an important prior for simplification of applying reinforcement learning to multiple agents[16]. Later the authors of [17] extended the same idea and successfully factored the joint function in terms of a linear combination of local functions and then used message-passing for finding joint optimal actions. In [18] the authors distributed the value function in terms of learning over multiple tables. However, they failed to extend the concept to the stochastic environments. In spite of this, policy-search techniques have found better success in partially observable environments [19]. In [20] the authors studied gradient-based distributed policy-search techniques.

The amount of work on multi-agent RL focused on the continuous action domain is less than that focused on discrete action domain. Some important approaches include [21] whose authors use discretization, and also [22] whose authors use Gaussian Networks for function approximation of the continuous action policy. Many of the previously mentioned techniques work only in inn restricted environments and also fail to extend to high dimensional input or to continuous actions. Also, the complexity of computation grows in exponential terms with the increase in the number of agents. Multi-agent control has been researched extensively in terms of the dynamics in problems like coverage control [23], formation control [24], and consensus [25]. For dynamical system approach the limitations lie in the requirements of the same for hand-engineered controls and use of domain-specific features. In spite of the fact that this technique allows developing provable characteristics about its controller, however it requires in-depth knowledge of the domain and also requires considerable hand-engineering. From a holistic perspective, deep reinforcement learning gives a more general approach for solving multi-agent problems sans the need for hand-engineering of features and other heuristics. It does so because the deep network is able to learn the properties of the controller without preprocessing of the raw input and rewards. In [26] the authors extend DQNs to train multiple agents independently. They show that collaborative and competitive behaviors arise with a proper choice of rewards. In [27] and [28] the authors train multiple agents for learning a communication protocol for solving shared utility tasks. They also show end-to-end training with the use of novel neural network architectures. However, inter-agent communication is a complex process whereas each agent has its own opinion and observations. Also use of novel neural network architectures is concurrent with advances in deep learning [29-31,6,7,32-35].

Recently, in order to deal with non-stationarity issues in Multi-agent Systems (MAS), Palmer et al. [36] have developed a technique named lenient-DQN (LDQN) wich applies leniency with decaying temperature values for adjustment of policy updates which are sampled from the experience replay pool. This technique has been applied to the coordinated multi-agent systems and the performance of this work has been compared with the hysteretic-DQN (HDQN) [37]. Experimentation demonstrates that LDQN is better than HDQN. The concept of leniency coupled with a replay strategy was also used in the weighted double deep Q-Network (WDDQN) in [38] in order to deal with non-stationarity. Experimentation shows that WDDQN performs better than double DQN in two multi-agent environments. Hong et al. [39] have introduced a deep policy inference Q-network (DPIQN) for modeling of MAS and also the enhanced version viz. deep recurrent policy inference Q-network (DRPIQN) for coping with partial observability. Experimentation shows generally better performance of both DPIQN and DRPIQN over their baselines viz. DQN and DRQN [40]. However, we do not use complex networks for our experimentation. It must be noted that our approach can be extended to any type of DQN including the state of art, as it involves using multiple similar DQNs as individual agents. To not exaggerate, we have limited our experiments to DQN and DDQN.

Gupta et al. [16] have examined three separate training techniques for an MAS i.e. centralized learning, concurrent learning and parameter sharing. Centralized policy gives a joint action using joint observations of all the agents in the environment while as the concurrent learning trains agents together with the help of the joint reward signal. In concurrent learning, each agent learns its private policy which is independent and is based on private observation. In the parameter sharing scheme, agents are trained simultaneously using their experiences together although each of them can have unique observations. Our scheme is not as complex as these. It involves sharing a common state and reward, with the exception of the action which is locally updated in the Experience Replay Pool of each agent. As a result, significant time and resources are saved during training and execution, because of the reduced complexity. Also, there is no inter-agent communication which is in scope for future work in this line of research.

## 3. Implementation and Experimentation

The proposed technique uses shared state and single common reward, but with agent-specific action updation for experience replay pooling of the DQNs. Each agent in the MAS is a Deep Q-Network. We use one DQN per action. Hence each DQN agent can have action values 0 or 1 which correspond to no action or action taken by the agent respectively. After all DQNs have predicted their action values, a decision structure is used to invoke final action whether it is in range (0 to 3) for LunarLander-v2 and Maze Traversal, or simply in range (0 to 1) for CartPole-v1. Note that for CartPole-v1, the action space for the proposed two agent system has four values (*agent_left:* **no-action, move-left**; *agent_right:* **no-action, move-right**), while as the environment action space has two values (**move-left, move-right**). Figure 1 shows the overall scheme for the proposed approach.

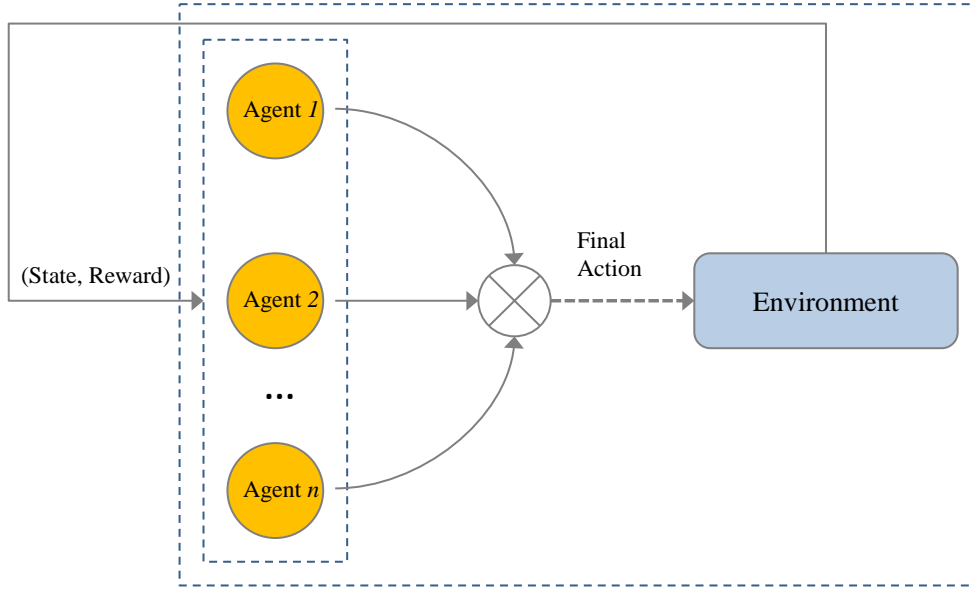

**Figure 1.** Overall scheme for the proposed approach

Experimentation was done on the Python platform using TensorFlow by an Intel® Xeon® (Two Core) processor, with 12 GB RAM and 16GB GPU. Since the proposed technique can be applied to any type of DQN, we do not test longitudinally along genres of Deep Q-Networks (DQNs), but instead investigate laterally in specific genres viz. DQN [1] and Double DQN (DDQN) [2]. The experiments are based on three tasks. These include two Open AI Gym tasks viz. Cartpole-v1 and LunarLander-v2, and a maze traversal task implemented locally. For testing we use the DQN and the DDQN. The code for the original tasks using conventional DQNs was borrowed from the links given in Table 1.

Table 1. Online Resources Used

| Task | Link |
|---|---|
| **CartPole-v1 DQN** | *https://colab.research.google.com/github/ehennis/ReinforcementLearning/blob/master/05-DQN.ipynb* |
| **CartPole-v1 DDQN** | *https://colab.research.google.com/github/ehennis/ReinforcementLearning/blob/master/06-DDQN.ipynb* |
| **Maze Traversal DQN** | *https://www.samyzaf.com/ML/rl/qmaze.html* |
| **LunarLander-v2 DQN** | *https://colab.research.google.com/github/davidrpugh/stochastic-expatriate-descent/blob/2020-04-03-deep-q-networks/_notebooks/2020-04-03-deep-q-networks.ipynb* |

## 3.1 Simulation Environments

### 3.1.1 CartPole-v1

In this OpenAI Gym Environment, a pole is attached to a cart by an un-actuated joint, which is moving along a frictionless track. The whole system is controlled by application of a force of +1 or -1 to the cart. The pendulum initially is upright, and the goal of the governing algorithm is to prevent the pendulum from falling over. The reward is +1 for each time step for which the pole remains upright. The episode ends if the standing pole is more than 15° from the vertical, or if the cart moves beyond 2.4 units from the center. The available actions are move left (0) and move right (1). The above environment is based on the cart-pole problem discussed by Barto, Sutton, and Anderson [41].

### 3.1.2 LunarLander-v2

In this OpenAI Gym Environment, the landing pad is located at coordinates (0,0). Coordinates correspond to the first two numbers in state vector. The reward for landing safely on the landing pad with zero speed is about 100 to 140 points. The episode ends if the lander crashes or comes to rest. These actions lead to additional -100 or +100 points respectively. Each leg ground contact gives +10. Firing main engine leads to -0.3 points each frame. Solved task has 200 points. There is infinite fuel in the lander, so it can learn to fly and then land on its first attempt. Four discrete actions are available i.e. do nothing (0), fire left orientation engine (1), fire main engine (2) and fire right orientation engine (3).

### 3.1.3 Maze Traversal

This is a customized framework for a Markov Decision Process (MDP) and consists of an environment (Maze with 8x8 cell structure) and an agent which acts in the environment. The environment is a square maze with three types of cells i.e. **o**ccupied cells, free cells, and the target Cell. The agent is self-maneuverable entity which is allowed to move only on free cells, and whose goal is to get to the target cell. The agent is encouraged to find the shortest path to the target cell by a simple reward scheme. Four discrete actions are available i.e. move left (0), move up (1), move right (2) and move down (3). The rewards are floating points ranging from -1.0 to 1.0. A move from a cell to another will be rewarded by a positive or a negative amount. A move from a cell to an adjacent one will cost the agent -0.04 points. This discourages the agent from wandering around and encourages it to get to the target in the shortest route possible. Maximum reward of 1.0 points is given when the agent arrives at the target cell. Entering a blocked cell costs the agent -0.75 points. An attempt to move to a blocked cell is invalid and is not executed. However, this attempt will incur a -0.75 points penalty. The same rule holds for an attempt to move outside the maze boundaries i.e. -0.8 points penalty. The agent is also penalized by -0.25 points for a move to a cell which it has already visited. In order to avoid infinite loops and wandering, the game ends if the total reward of the agent is below a negative threshold i.e. -0.5 x

Maze-Size. It is assumed that under this threshold, the agent has lost its way and has made too many errors from which it has learned enough, and should proceed to a new episode/game.

Figure 2 shows rendering screenshots for the three tasks.

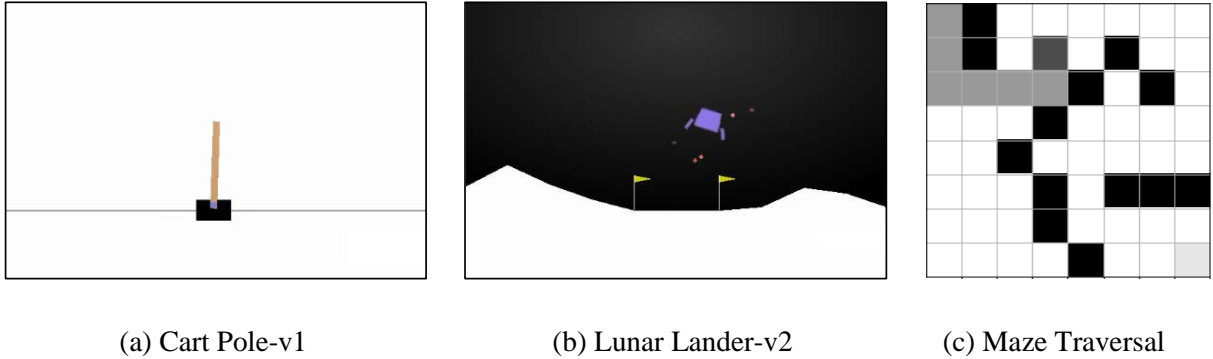

(a) Cart Pole-v1  (b) Lunar Lander-v2  (c) Maze Traversal

**Figure 2.** Screenshots of the three tasks

## 3.2 Results

The results for various environments are discussed in this section.

### 3.2.1 Cartpole-v1

Figure 3 shows the *reward* v/s *episode* variation for Cartpole-v1 for DQN and Class Specific DQN approach (CS-DQN) i.e. the proposed technique. The red horizontal line corresponds to reward for solving the task i.e. reward=195. The dark brown and light brown vertical lines correspond to the episode number for which the task is solved by CS-DQN and DQN respectively.

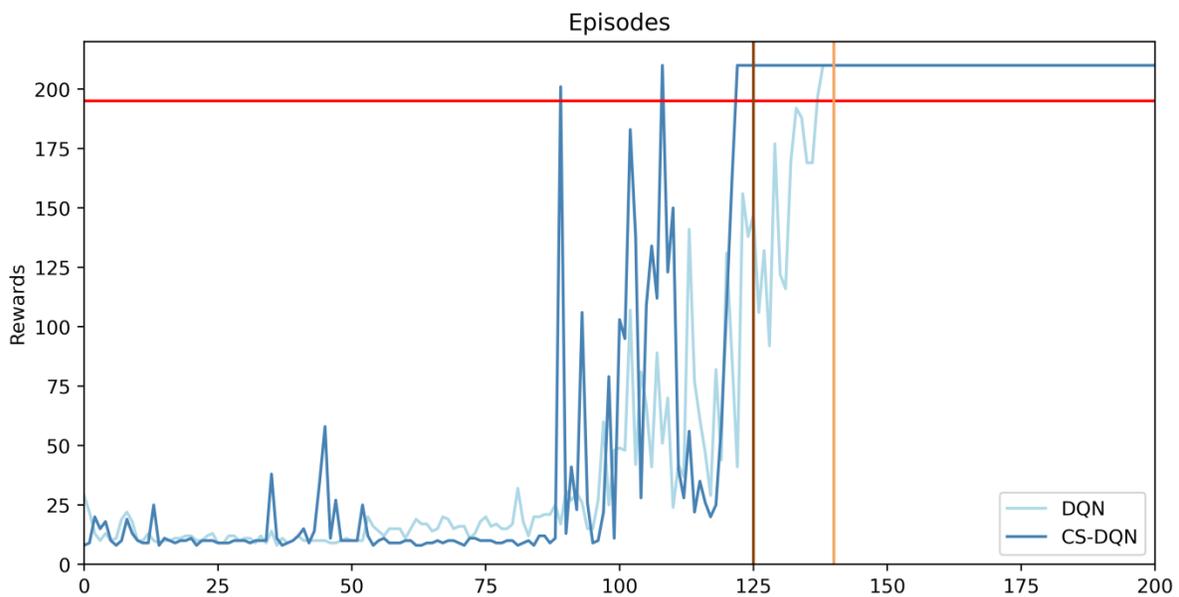

**Figure 3.** Rewards v/s Episodes obtained by using DQN with conventional and proposed approaches for Cartpole-v1respectively

Figure 4 shows the results within same environment and set of rules but by using DDQN for conventional and proposed approaches. The proposed approach is referred to here as CS-DDQN (Class Specific DDQN). The horizontal red, vertical light-brown and vertical dark-brown lines have the same meanings as the previous figure.

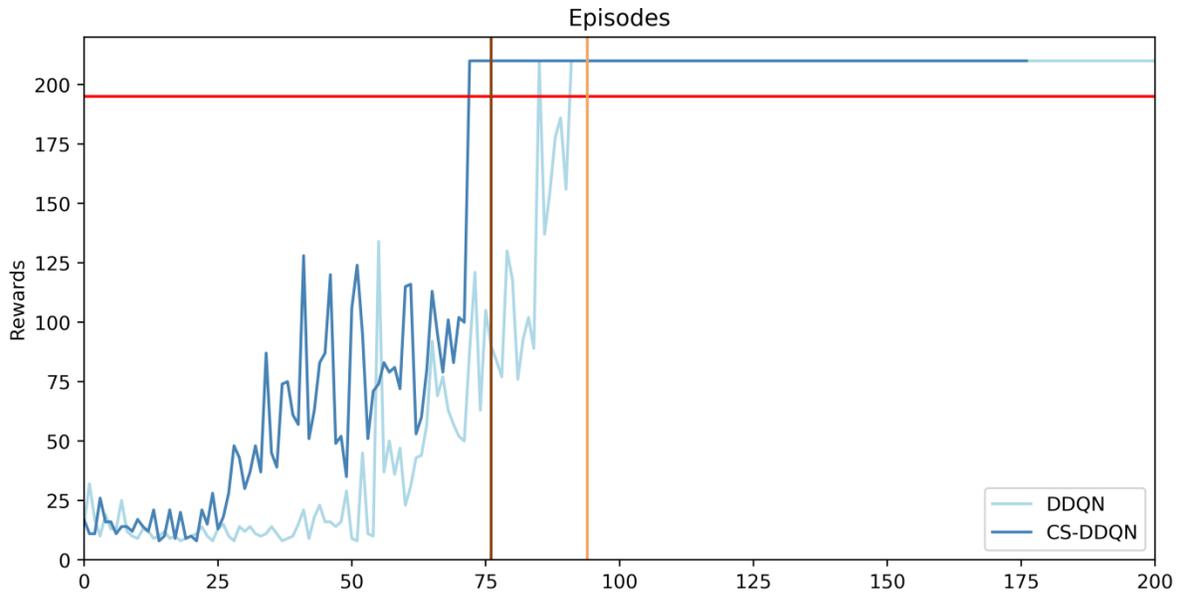

**Figure 4.** Rewards v/s Episodes obtained by using DDQN with conventional and proposed approaches for Cartpole-v1respectively

It is noted from Figures 3 and 4 that the proposed approach converges much earlier as compared to the conventional approach. After training, they are allowed to run to evaluate their respective performances, which stay well above the minimum reward line.

### 3.2.2 LunarLander-v2

Figure 5 shows the *episode score* v/s *episodes* for DQN and CS-DQN on the LunarLander-v2 task. It should be noted that CS-DQN is able to outperform the conventional DQN based approach by converging in 671 episodes, whereas DQN lags behind by converging in 749 episodes. The target score for the task is 200 points which is showb by the horizontal dotted line.

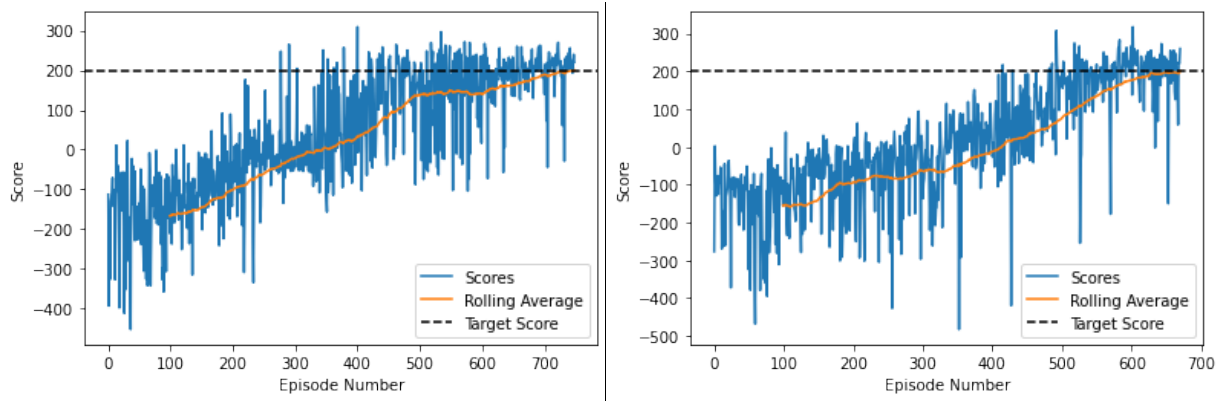

**(a) DQN** (Task completed in 749 Episodes)  **(b) CS-DQN** (Task completed in 671 Episodes)

**Figure 5.** Score v/s Episode performance for DQN and CS-DQN on the LunarLander-v2 task

### 3.2.3 Maze Traversal

For the maze traversal task, Figure 6 shows the *episode length per epoch* v/s *epochs* for DQN and CS-DQN approaches respectively. The lesser the episode length, the lesser is the number of steps which the agent takes to reach the target cell for the particular epoch. The training stops when the *win-rate* becomes 1.

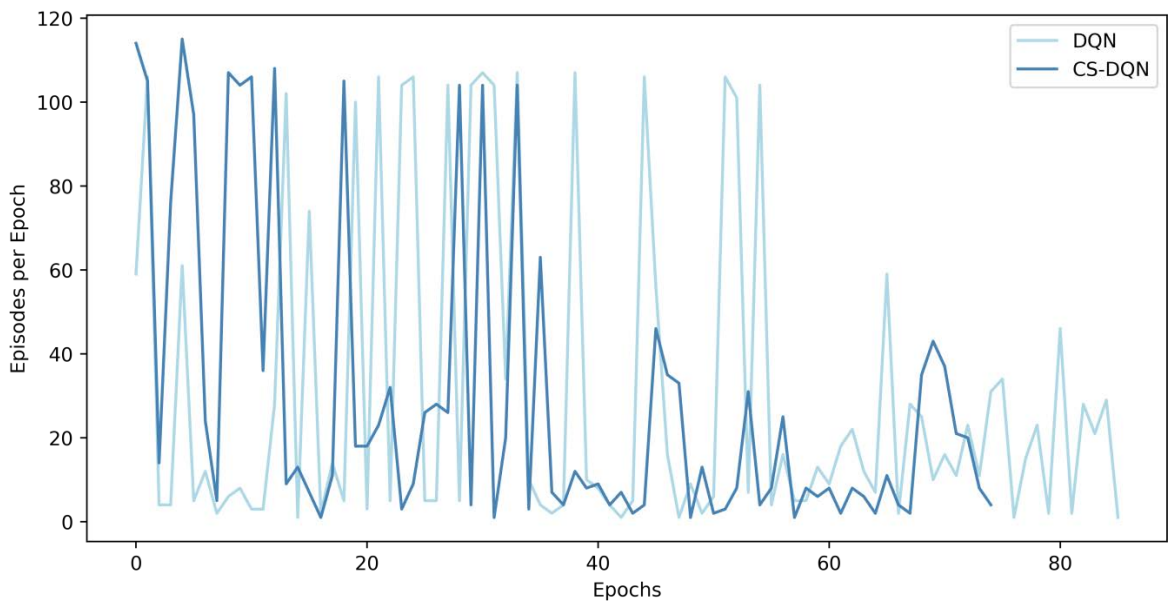

**Figure 6.** Episode length v/s epochs for Maze Traversal Task

It should be noted that the proposed approach converges much before the conventional approach i.e. in lesser number of epochs. Convergence takes place in this task when w*in-rate* becomes 1. Figure 7 shows the total win count for both approaches till w*in-rate* becomes 1. Note that the total win count towards the end of training is higher for CS-DQN as compared to DQN.

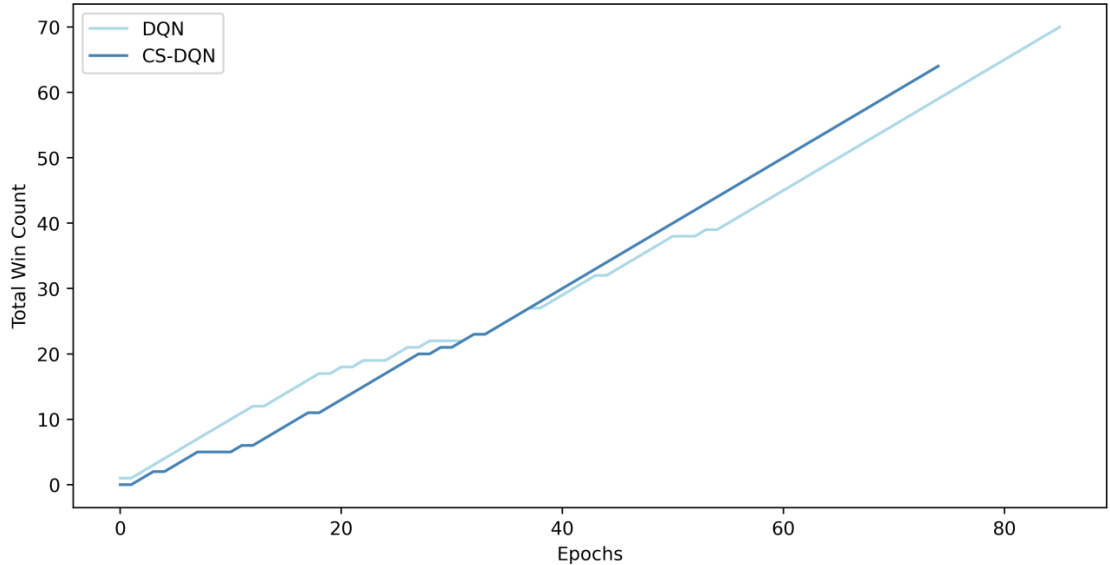

**Figure 7.** Total win count v/s epochs for the Maze Traversal task

As is noted from various experiments, our proposed approach outperforms the conventional approaches on various tasks used in this work.

## 4. Conclusion

In this paper, the importance of keeping simplicity in place in the scheme of DQN based Multi-agent Systems (MAS) for Reinforcement Learning (RL) was highlighted, in light of difficulties faced by the contemporary systems in training, need for resources, fine-tuning, performance, etc. Consequently, a simple but efficient scheme is proposed which consists of using DQN agents with binary actions for MAS based RL. The proposed approach can be used with any DQN, hence we use plain DQN and DDQN (Double Q-learning). It should be noted that we do not compare our approach with MAS for RL because our system is much simpler and is not an extension of MAS for RL as per various MAS schemes. The proposed system is tested on two Open AI Gym Environments i.e. *CartPole-v1* and *LunarLander-v2*, and one customized environment i.e. *Maze Traversal*. The proposed system is able to outperform the baseline on these tasks by a decent margin. Future work would involve using communication between agents and increasing the efficiency of the MAS.